\documentclass{article}

     \usepackage[final, nonatbib]{neurips_2020}

\usepackage[utf8]{inputenc} % allow utf-8 input
\usepackage[T1]{fontenc}    % use 8-bit T1 fonts
\usepackage{hyperref}       % hyperlinks
\usepackage{url}            % simple URL typesetting
\usepackage{booktabs}       % professional-quality tables
\usepackage{amsfonts}       % blackboard math symbols
\usepackage{nicefrac}       % compact symbols for 1/2, etc.
\usepackage{microtype}      % microtypography

\usepackage{hyperref}       % hyperlinks
\usepackage{url}            % simple URL typesetting
\usepackage{amsfonts}       % blackboard math symbols
\usepackage{nicefrac}       % compact symbols for 1/2, etc.
\usepackage{amsthm}
\usepackage{comment}
\usepackage{float}
\usepackage{enumitem}
\usepackage{array}
\usepackage{wrapfig}

\usepackage{color}
\usepackage{caption}

\usepackage{boxedminipage}
\usepackage{latexsym}
\usepackage{amsfonts,amsmath,algorithm}
\usepackage{amssymb}
\usepackage{bm}
\usepackage{mathabx}

\usepackage{graphicx}
\usepackage{caption}
\usepackage{float}
\usepackage{subcaption}
\usepackage{wrapfig}
\usepackage{xspace}
\usepackage{multirow}
\usepackage{mathtools}
\usepackage{physics}

\RequirePackage[textsize=scriptsize]{todonotes}

\usepackage{amsmath}
\usepackage{amsthm}
\usepackage{amssymb}
\usepackage{bbm}
\usepackage{mathtools}
\usepackage{mathrsfs}
\usepackage{dsfont}
\mathtoolsset{showonlyrefs}

\usepackage{courier} % For \texttt{foo} to put foo in Courier (for code / variables)

\usepackage{lipsum} % For dummy text

\usepackage[space]{grffile} % For spaces in image names

\usepackage[round]{natbib}

\usepackage{hyperref}

\usepackage{graphicx}
\usepackage{wrapfig}
\usepackage{xcolor}
\definecolor{dark-red}{rgb}{0.4,0.15,0.15}
\definecolor{dark-blue}{rgb}{0,0,0.7}
\hypersetup{
    colorlinks, linkcolor={dark-blue},
    citecolor={dark-blue}, urlcolor={dark-blue}
}

\usepackage{algpseudocode}
\usepackage[vlined,linesnumbered,ruled,algo2e]{algorithm2e}
\SetKwProg{Fn}{Function}{}{}
\let\oldnl\nl% Store \nl in \oldnl
\newcommand{\nonl}{\renewcommand{\nl}{\let\nl\oldnl}}% Remove line number for one line

\usepackage{multicol}
\usepackage{enumitem}
\usepackage[utf8]{inputenc} % allow utf-8 input
\usepackage[T1]{fontenc}    % use 8-bit T1 fonts
\usepackage{hyperref}       % hyperlinks
\usepackage{url}            % simple URL typesetting
\usepackage{booktabs}       % professional-quality tables
\usepackage{nicefrac}       % compact symbols for 1/2, etc.

\usepackage{theoremref}

\NewDocumentCommand{\tens}{e{_^}}{%
  \mathbin{\mathop{\otimes}\displaylimits
    \IfValueT{#1}{_{#1}}
    \IfValueT{#2}{^{#2}}
  }%
}

\title{Bosonic Random Walk Networks for Graph Learning}

\author{%
  Shiv Shankar\\ 
  College of Information and Computer Science\\
  University of Massachusetts\\
  \texttt{sshankar@cs.umass.edu} \\
  \And
  Don Towsley \\
  College of Information and Computer Science\\
  University of Massachusetts\\
  \texttt{towsley@cs.umass.edu} \\
}

\begin{document}

\maketitle

\begin{abstract}
%The last two decades have development of many techniques towards understanding and using graph-structured data. The same period has also seen tremendous progress in quantum computing. These breakthroughs have led to new algorithms and techniques which are more efficient than classical algorithms. In this project we intend to explore application of such quantum techniques into graph learning problems. Building upon recent works in quantum computation we explore two different quantum inspired techniques: a) based on bosonic walks and b) based on gaussian boson sampling; within the context of improving learning on graphs. 

The development of Graph Neural Networks (GNNs) has led to great progress in machine learning on graph-structured data. These networks operate via diffusing information across the graph nodes while capturing the structure of the graph. Recently there has also seen tremendous progress in quantum computing techniques. In this work, we explore applications of multi-particle quantum walks on diffusing information across graphs.
Our model is based on learning the operators that govern the dynamics of quantum random walkers on graphs. We demonstrate the effectiveness of our method on classification and regression tasks.
 
%We present diffusion-convolutional neural networks (DCNNs), a new model for graph-structured data. Through the introduction of a diffusion-convolution operation, we show how diffusion-based representations can be learned from graph-structured data and used as an effective basis for node classification. DCNNs have several attractive qualities, including a latent representation for graphical data that is invariant under isomorphism, as well as polynomial-time prediction and learning that can be represented as tensor operations and efficiently implemented on the GPU. Through several experiments with real structured datasets, we demonstrate that DCNNs are able to outperform probabilistic relational models and kernel-on-graph methods at relational node classification tasks. 

%on graph-structured data has made great improvements since ve proven effective in many domains. These graph neural networks often diffuse information using the spatial structure of the graph. We propose a quantum walk neural network that learns a diffusion operation that is not only dependent on the geometry of the graph but also on the features of the nodes and the learning task. A quantum walk neural network is based on learning the coin operators that determine the behavior of quantum random walks, the quantum parallel to classical random walks. We demonstrate the effectiveness of our method on multiple classification and regression tasks at both node and graph levels.

\end{abstract}

\section{Introduction}
The current era of ubiquitous connectivity has provided researchers with ever-increasing troves of data. Most of such ‘real-world’ data have an underlying graphical structure that can be utilized to build better models and derive greater insights. %Graphs have been a common abstraction to understand and categorize these structures.
Such graphical structures are not limited to the web, social networks, or other network systems. Graph-structured problems are also common in many scientific fields such as immunology \citep{Crossman2020}, chemical analysis \citep{PCSJohn2019} and bio-chemistry \citep{bonetta2019}.

Current machine learning approaches for analyzing structured data can be broadly categorized into neural approaches and classical approaches. Classical approaches rely on comparing graphs by utilizing various similarity notions \cite{kondor09_graphlet,kondor08_spectrum}. Similarity comparison between graphs can be performed directly via walks \citep{dobson03_enzyme,callut_dsicriminativewalk}. Another related technique is to use graph kernels \citep{JMLR:v11:vishwanathan10a,gartner03_graphhard}. There have been some recent works \citep{bai17_qkernel} that try to define similarity using quantum walks. 

The last decade also saw great progress in machine learning via the development of deep-learning techniques. Some of these works also focused on applying neural networks to graph-structured data \citep{Defferrard2016ConvolutionalNN,NIPS2015_5954} 
\citet{NIPS2015_5954} present a method for differentiable fingerprinting where the hashing functions are replaced by neural networks. \citet{Defferrard2016ConvolutionalNN} extend the convolution operator to graphs using graph Laplacians. \cite{Kipf2017SemiSupervisedCW} use the same technique for semi-supervised learning on graphs.
\cite{Atwood2016DiffusionConvolutionalNN} also, extend convolutions to graphs via graph diffusions. Building upon these works and the ideas of \citet{bai17_qkernel}, \citet{Dernbach19_qwnn} incorporate quantum walks into a neural network.

In this paper, we explore the application of some quantum computing techniques in graph learning. We first summarize some basic principles relevant to our approaches in Section \ref{sec:prelim}. Next, we present a graph learning method that is inspired by these quantum ideas. Our approach is a hybrid one that a) uses quantum walks to learn diffusions  and b) utilizes the diffusions in a classical way. Finally, we present the results of our experiments.

\section{Preliminaries}
\label{sec:prelim}
%In this section we provide an overview of quantum walks and bosonic quantum principles.

\subsection{Bosonic Quantum Mechanics}
Photonic circuits are a prime candidate for both near-term and future quantum devices. Photons are a type of boson that lead to interesting statistical and physical phenomena. Hence understanding some key aspects of bosonic quantum systems is important when considering possibly physical realization of some quantum algorithms. 

%Bosons are one of the two classes of fundamental particles. \footnote{ These are named after SN Bose who developed the theory and characteristics of these family of particles which were at that time unknown}
%Another is that there is no restriction on the number of them that occupy the same quantum state. 

%Examples of bosons are fundamental particles such as photons, gluons, and W and Z bosons (the four force-carrying gauge bosons of the Standard Model), the recently discovered Higgs boson, and the hypothetical graviton of quantum gravity. Some composite particles are also bosons, such as mesons and stable nuclei of even mass number such as deuterium (with one proton and one neutron, atomic mass number = 2), helium-4, and lead-208;[a] as well as some quasiparticles (e.g. Cooper pairs, plasmons, and phonons).[9]:130

An important characteristic of bosons is that two bosons of the same type are indistinguishable; and this has interesting consequences. 
Let us denote system state by $\ket{\psi(x_1,x_2)}$ where $\psi$ is some function and $x_1,x_2$ are generalized coordinates (eg position, state etc) of the two particles. The exchange operator $E$ is then defined by the following action

$$
E\ket{\psi(x_1,x_2)} = \ket{\psi(x_2,x_1)}
$$

Informally the exchange operator swaps the coordinates/states of the individual particles in the combined system.
The requirement of indistinguishability is the invariance of a multi-particle system to the exchange operator. 
%his implies that only those states which are unaffected by the exchange operator are physically possible. This is best described by the following example. 
Imagine a system having binary states $\ket{0},\ket{1}$ and two particles that can occupy those states. The standard computational decomposition for the Hilbert space of such a system has 4 basis viz. $\ket{00},\ket{11},\ket{01},\ket{10}$; corresponding to the states of each of the two particles. One can imagine that the system is described by the state $\ket{01}$. However, indistinguishability means we can permute the labels of the particles and the system is equally be described by $\ket{10}$. Note that this along with the superposition principle implies that any superposition of the these states is an equally valid description of the state. This leads to the following principle:

\paragraph{Symmetrization Postulate for Bosons} In a system of indistinguishable bosons, the only possible states of the system are ones that are symmetric with respect to permutations of the labels of those particles.

The Symmetrization Postulate restricts the Hilbert space of the system to lie in the completely symmetric subspace. In the specific case considered above the Hilbert space of this system is spanned by 3 states instead of the usual 4. Furthermore the postulate implies that result of any measurement of a state must project the state into the symmetric indistinguishable subspace.

%Note that the last basis seems like an entangled state. While indeed it is an entangled state for the standard 2 qubit system, it is not so for the system of 2 indistinguishable bosons \cite{Sciara_2017}. Hence instead of usual entanglement there can be 'quasi-entanglement' effects.

\subsubsection{Fock space}
Indistinguishability also implies that only the total number of bosons in a given state has any meaning. This makes it convenient to use an alternate basis for describing the Hilbert space of the system known as the Fock basis. It is a construction for the state space of a variable or unknown number of identical particles from the Hilbert space of a single particle. For bosons, the $n$-particle states are elements in the symmetric product of $n$ uni-particle Hilbert spaces. For more details of the Fock space notation refer to the Appendix \ref{apx:fock}

\paragraph{Creation and Annihilation Operators}
Operations in the Fock space bases are written as unitary matrices of creation and annihilation operators. As its name suggests the creation (or raising) operator (denoted commonly by $a$) adds a particle to the state it operates on, while the annihilation (lowering) operator ($a^\dagger$) does the opposite. \\

\subsection{Quantum Walks}
A classical walk on a graph $G=(V,E)$ can be described as follows. At any given timestep $t$ the walker resides at one of the nodes of the graph say $u$. Each node has an associated multiheaded coin with each head corresponding to an incident edge. Equivalently there is a multinomial distribution over the edges incident on the node. The walker then draws an edge according to the distribution (say $e = (u,v)$). The walker then moves to the node $v$ which is connected to $e$.

Quantum walks are the quantum extension of a classical random walk.
A classical walk involves a walker moving around on a graph and at any point in time its position is given by a probability distribution. A quantum walk is similar, however instead of a random process the walkers movement is governed by a sequence of unitary operations. Unlike a classical walk where the walker can only be at a node, the quantum walker can be in a superposition over all nodes in the graph. 

%\subsection{Classical and Quantum Walks}
%

We follow the approach of \cite{kendon06_walks} in describing quantum walks. A quantum walk involves two Hilbert spaces: the position space $\mathcal{H}_v$ corresponding to nodes of the graph; and a coin space $\mathcal{H}_c$. To preserve unitarity the size of the coin space is fixed across all nodes. This can be achieved by taking the maximum degree of the nodes as the size of the space.
The quantum walker's state is determined by the combined space of position and coin combinations. Instead of a coin toss, we now have a unitary operator $C$ (called the coin operator) on the coin space describing the evolution of the coin-part of the walkers state. We also have a shift operator $S$. The shift operator acts as a conditional gate: depending on the coin state it swaps the coefficients of the corresponding positions. The evolution of the entire system is given by the unitary operator $U = S(I\tens C)$

\begin{figure*}
    \centering
    \begin{subfigure}{0.45\linewidth}
    \includegraphics[width=\linewidth]{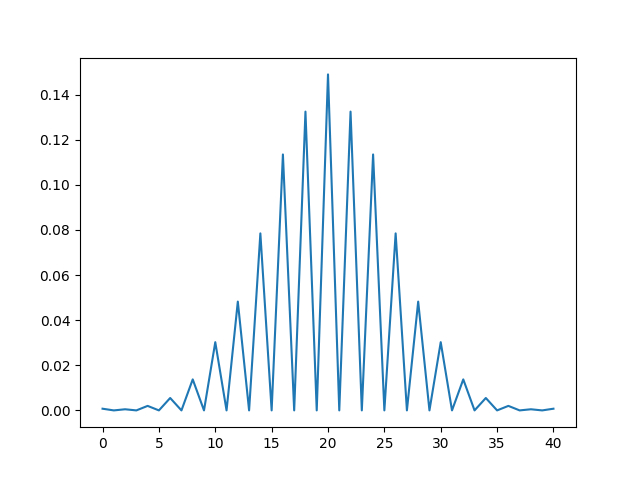}
    \caption{Classical Random Walk  \label{fig:cvsq_walk:a} }
    \end{subfigure}\qquad
    \begin{subfigure}{0.45\linewidth}
    \includegraphics[width=\linewidth]{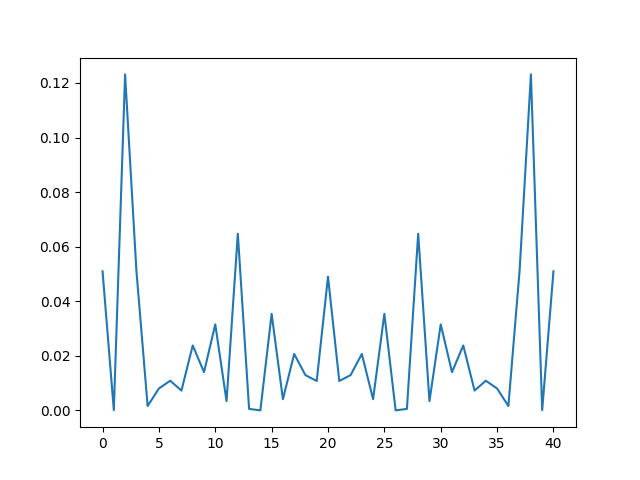}
    \caption{Quantum Random Walk  \label{fig:cvsq_walk:b} }
    \end{subfigure}
    \caption{Comparing classical and quantum walk \label{fig:cvsq_walk}}
\end{figure*}

This kind of evolution produces a behavior completely unlike that of a classical walk. The superposition of states allows walker trajectories to interfere: something that cannot happen classically. This interference can lead to a faster spreading of the walker's final position distribution. The clear effect of this can be seen in Figure \ref{fig:cvsq_walk}. These figures show the result of simulating a classical and quantum walk on a 1-d lattice for 30 steps.
The final state distribution of classical walk is shown in Figure \ref{fig:cvsq_walk:a}. It is clearly centered around its starting point and has a exponentially falling tail as one moves further from the start.

The quantum walk however ( Figure \ref{fig:cvsq_walk:b}) shows a very different picture. While the mean of the walk is still at the starting point, the distribution modes are peaks far away from the start. Such behavior allows a quantum random walker to have significantly better exploration. Inspiring from this insight \cite{Dernbach2018QuantumWI} proposed a version of diffusion networks based upon quantum walks. They demonstrated that using probability distribution of quantum walkers allows for far better exploration and incorporation of graph structure as compared to classical approaches.

\section{Bosonic Walks Networks}

A natural question with respect to random walk based graph networks is whether incorporation of multiple walkers can lead to a different outcome. While multiple non-interacting classical walkers have no extra power compared to a single classical walker, the answer is different for quantum walkers \citep{Chandrashekar_2012}. A key reason for this is the symmetrization postulate referred to earlier.
%For an illustrative example refer to Appendix \ref{apx:example}. 

\subsection{Bosonic Quantum Walks}
Bosonic walks can a) have unintuitive non-local correlations across walker states and b) allow for dynamics not accessible for distinguishable particles. As such even limited bosonic walks can have surprising power. For example \cite{gamble2010twoparticle} demonstrate that there are classes of non-isomorphic graphs that can be distinguished by the node distribution of multi-particle walks but not by the node distribution of a single particle quantum walk.
More recently \cite{Lahini_2018} have proposed a scheme for implementing high-fidelity quantum gates using a multiple bosonic quantum walkers. 
%In fact we will see this comparatively limited space allows for diverse forms of behavior once we consider indistinguishable walker. 

%%However we will abstract out the coins as it makes it easier to work with multiple walkers. Technically we are looking at that subset of quantum walks where the coin space is constrained to be independent. Recall that 'quasi-entangled' are natural in the Hilbert space of multiparticle bosonic systems. 

%Specifically we will focus on quantum walks of bosonic walkers.
We focus on a setting with multiple bosonic particles executing quantum walks on a graph. The Hilbert space of these walkers allow significantly more trajectories of the walker by allowing entangled coins and other complexities. The evolution of the walkers state in the Fock space basis is driven by the Hamiltonian $H$ given by

$$ H = A_{ij}c_i^\dagger c_j + E c_ic_i^\dagger (c_ic_i^\dagger- 1)$$

$A$ is the adjacency matrix of the graph and $c_i,c_i^\dagger$ are creation and annihilation operators associated with node $i$. 
%Ignoring $U$ for the moment it is easy to see how the action of $A_{ij}c_i^\dagger c_j$ moves the walker from state $i$ to state $j$.
$Ec_ic_i^\dagger (c_ic_i^\dagger- 1)$ is a term which describes the interaction between the walkers and $E$ is the interaction strength. Note that when the particles are in different nodes, the interaction term has no effect. % by the interaction term. One might imagine that if the interaction strength is 0 , the system behaves same as two independent walkers. As it turns out this is not the case. This is partly because of the 'quasi-entanglement' effects even among non interacting particles mentioned earlier.

\begin{figure*}
    \centering
    \begin{subfigure}{0.45\linewidth}
    \includegraphics[width=\linewidth]{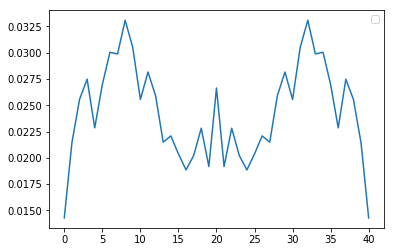}
    \caption{Quantum Bosonic Random Walk E=0  \label{fig:qwb} }
    \end{subfigure}\qquad
    \begin{subfigure}{0.45\linewidth}
    \includegraphics[width=\linewidth]{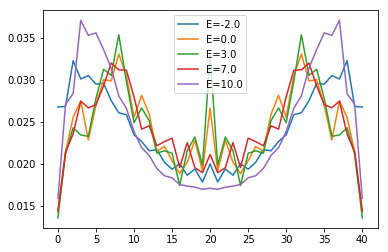}
    \caption{Quantum Bosonic Random Walk for varying E \label{fig:qwbr} }
    \end{subfigure}
    %\qquad
    %\begin{subfigure}{0.45\linewidth}
    %\includegraphics[width=\linewidth]{figs/qrw2_i100.png}
    %\caption{Quantum Bosonic Random Walk Fig(b) + U=100  %\label{fig:qwball} }
    %\end{subfigure}
    \caption{Comparing quantum walks with difference interaction energy $U$ \label{fig:qwb2}}
\end{figure*}

Similar to earlier results, we compute the probability of observing a walker across different nodes on a 1-D lattice after 30 steps in the case when $E=0$. This is plotted in Figure \ref{fig:qwb}, where we can notice easy differences between this plot and the plot of the single particle quantum walk discussed earlier. Next we plot the probability of observing a walker for different values of $E$ in Figure \ref{fig:qwbr}. Note the probability distribution changes significantly as the interaction strength changes. %One also observes that with a strong enough interaction term the probability distribution peaks at the start (similar to a classical walk) \ref{fig:qwball}.

%Two kinds of quantum walks have been introduced in the literature; namely, continuous time quantum walks (Farhi and Gutmann 1998; Rossi et al. 2017) and discrete time quantum walks (Lovett et al. 2010). Quantum walks have recently

\subsection{Bosonic Quantum Walk Neural Networks}
%\cite{Dernbach2018QuantumWI} proposed quantum walk neural networks that rely on computing the probability distribution of walkers on a graph to aggregate information. 
A Bosonic Quantum Walker Network (QWB) is the natural extension of the \citet{Dernbach2018QuantumWI} model using bosonic quantum walkers. For our description below, we will follow a similar presentation.
The key idea behind a quantum walk neural network is to use the  walker's distribution over the nodes of a graph to construct a diffusion matrix, which is then utilized to aggregate information from the nodes.
At each time step,  we simulate the dynamics of the walker using the coin operator, $C$, to modify the spin state of the walkers $\psi$ according to $ C (t) \psi_{t} \xrightarrow{} \psi_{t+1} $. The coin operator need not be static and can depend on both time and node features. This is followed by the shift operator, which moves the walker to a neighbouring node depending upon the walker spin state. The walker dynamics induces a probability distribution of the walker over the graph (written as a probability matrix $P$). 
Next, this matrix $P$ is used to diffuse the node level features across the graph: $\hat{X} = PX$. These diffused features are the output of a single quantum diffusion layer. These features can then be used either as input for a second diffusion layer; or for final prediction. All of these operations are differentiable, and hence we can use backpropagation to compute the gradient of the loss with respect to all the model parameters (especially parameters of the coin matrix) . 

%For the bosonic quantum walks we need to perform operations only in the symmetric Hilber space. This can be achieved via evolving single particle walkers and then performing a projection into the indistinguishable space by performing repeating sums over identical labels. A different way to 

\begin{figure}
    \centering
    \includegraphics[width=\linewidth]{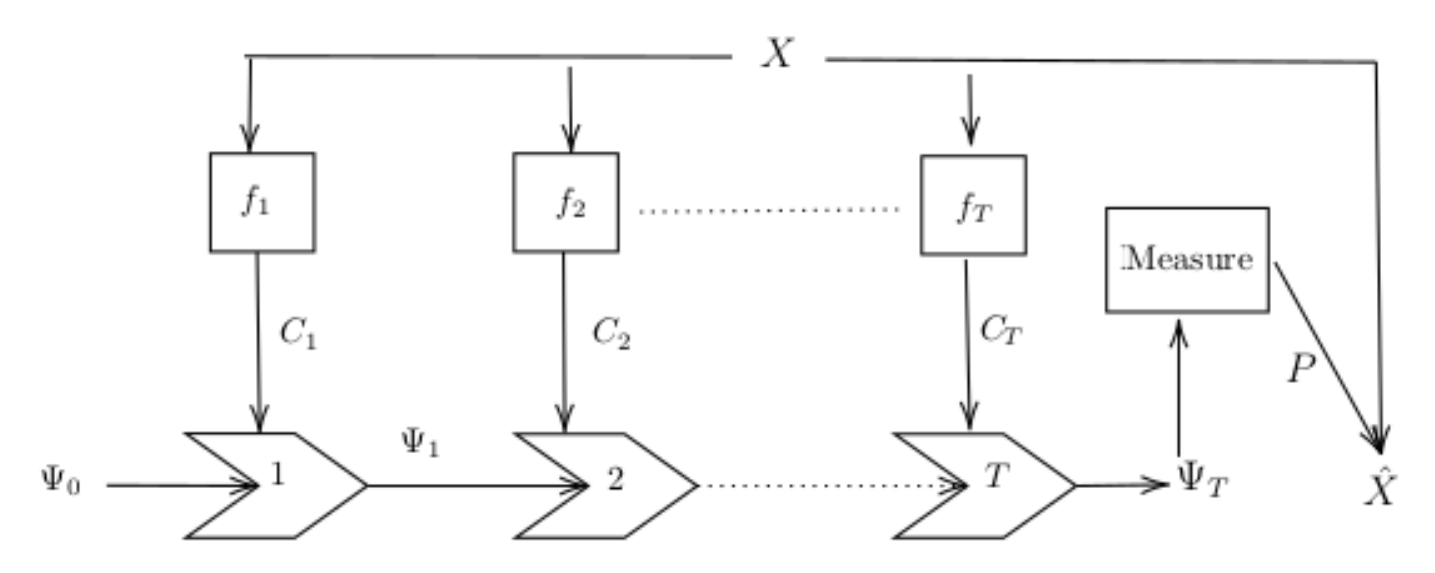}
    \caption{Quantum Walk Neural Network Schematic: The feature matrix $X$ is used to produce the coin operators $C_i$ used in each step $i$. The superposition $\psi$ evolves after each step. The final layer diffuses $X$ using measured probabilities $P$ to compute $\hat{X}$}
    \label{fig:qwnn}
\end{figure}

%\subsection{Bosonic Quantum Walk Neural Networks}
Note that the description till here is independent of how the walkers behave. In fact, walkers can behave completely classically, in which case the behaviour is identical to the Diffusion Convolution model of \citet{Atwood2016DiffusionConvolutionalNN}. For a quantum walk network, the walk dynamics are governed by quantum evolution. The induced probability matrix is the one determined by the measurement of the walker's node-state. We compute $ P $ induced by a bosonic multi-particle quantum walk on the graph in the bosonic quantum walk model. This can be achieved via evolving single-particle walkers and then performing a projection into the indistinguishable space by performing repeating sums over identical labels. More details are available in Appendix \ref{apx:bosonic} 

Since a $k$-particle quantum walk naturally produces a symmetric superposition over $k$-tuples of nodes, we can, in principle, extend $P$ from a distribution over nodes to a distribution over node pairs or even higher orders. In our experiments, however, we do not use such higher-order features. Instead, we compute $P$ from the probability of observing \emph{a single} boson at the given nodes.

\section{Experiments}
\label{sec:exp}
\paragraph{Datasets}
We experiment with commonly used graph datasets: QM7, which is a regression task and  MUTAG, NCI1, and Enzymes (which are classification problems). MUTAG \cite{denath_mutag} is a dataset of 188 mutagenic aromatic and heteroaromatic nitro compounds that are classified as either mutagenic or not. NCI1 \cite{wale_nci1} consists of 4110 graphs representing two sets of chemical compounds screened for activity against non-small cell lung cancer. For both these datasets, each graph represents a molecule, with nodes representing atoms and edges representing bonds between atoms. Each node has an associated label that corresponds to its atomic number. Enzymes \cite{borgwardt2005protein} is a dataset of 600 molecules where the task is to classify each enzyme into one of six classes.
The QM7 dataset \cite{PhysRevLett.108.058301,blum09} is a collection of 7165 molecules, each containing up to 23 atoms. The goal of the task is to predict the atomization energy of each molecule.

%\paragraph{Computational Challenges} Unlike classical graph neural networks these quantum inspired techniques exhibit terrible scaling properties. Simulating a $k$-particle quantum walk scales exponentially in $k$ as the state space of the system is a roughly the same size as the product of individual walker space. More specifically for a graph of size $V$ the complexity of the above simulation grows as $O(V^k)$ This implies that simulating even a 2-particle walk on a graph of 20 nodes takes roughly the same time as a single walk on 400 nodes. GBS on the other hand does not have a scaling issue with respect to number of photons. However computation of Permanents and Hafnians is also exponential in size of the matrices which are of the order of the size of the graphs. As such in our experiments we had to restricted ourselves to graphs of size < 20. 

\paragraph{Experimental Details}
We include as baselines two classical methods (DCNN, GCN). DCNN refers to the diffusion convolutional network of \citet{Atwood2016DiffusionConvolutionalNN}, while GCN is the graph convolution architecture of \citet{Kipf2017SemiSupervisedCW}.
QWNN is the quantum walk based model presented in \cite{Dernbach19_qwnn}. QWB2 is our two-particle bosonic extension of the QWNN model. The metric used for classification tasks (NCI, MUTAG, Enzymes) is accuracy (so higher the better) while the one used for QM7 is mean prediction error (so lower the better).

Classical simulation of quantum walks has poor scaling properties. Simulating a $k$-particle quantum walk scales exponentially in $k$. As such in our experiments we restricted ourselves to graphs of size less than 70 and had to reduce sizes of feature embeddings. Furthermore to keep comparison fair in terms of feature size between classical and quantum models, we applied the same restriction to the classical models as well. For more details on the parameters used refer to Appendix \ref{apx:hyper}.

\paragraph {Results}
Table \ref{tab:class_results} reports the classification performance of different models. We see that QWB2 outperforms other models, especially on Enzymes where the gain is substantial. QWB2 also outperforms QWNN and other classic approaches on MUTAG and NCI datasets.
Regression results on QM7 are presented in Table \ref{tab:reg_results}. The basic trend of QWB2 outperforming other approaches remains, thought the performance differences are comparatively smaller. 
Overall QWB2 seems to outperform all other models including its single-walker counterpart QWNN.

\iffalse{
\begin{table}[h]
    \centering
    \begin{tabular}{|l||c|c|c|}
    \hline
        Model & NCI1 & MUTAG & QM7 \\ \hline
        
        DCNN & 77.1 & 74.3 & \textbf{80.1} \\
        QWNN & 72.6 & 60.0 & 111.6\\
        QWB2 & \textbf{78.1} & 63.2 & 98.7\\
        GBS & 74.7& \textbf{78.7} & 100.7\\  \hline
    \end{tabular}
    \caption{Results of different models on graph classification and regression tasks}
    \label{tab:results}
\end{table}
}\fi

\begin{table}[h]
    \begin{subtable}{0.6\textwidth}
    \centering
    \begin{tabular}{|l||c|c|c|}
    \hline
        Model &  Enzymes & MUTAG & NCI   \\ \hline
        GCN & 31.4 & 87.4 & 69.6 \\
        DCNN & 27.9 & 89.1 & 69.1 \\
        QWNN & 33.6 & 88.4 & 73.6\\
        QWB2 & \textbf{40.2} & \textbf{90.0} & \textbf{76.7}\\ \hline
    \end{tabular}
    %GBS & 74.7& \textbf{78.7} & 100.7\\  \hline
    \caption{Classification tasks \label{tab:class_results}}
    \end{subtable}
    \quad
    \begin{subtable}{0.3\textwidth}
    \begin{tabular}{|l||c|c|}
    \hline
    Model & MSE & MAE \\ \hline
        GCN & 17.5 & 12.4 \\
        DCNN & 11.9 & 8.6 \\
        QWNN & 10.9 & 8.4 \\
        QWB2 & \textbf{9.2} & \textbf{7.9}\\ \hline
    
    \end{tabular}
    \caption{Atomization energy prediction on QM7 \label{tab:reg_results}}
    \end{subtable}
    \caption{Results of different models on graph classification and regression tasks}
    \label{tab:results}
\end{table}

\section{Related Work}

\paragraph{Graph Neural Networks}
Early graph neural networks (GNN) \citep{gori2005new, scarselli2008graph} used recursive architectures to encode graphs into finite-dimensional vectors. Since then has been tremendous progress in learning representations of graphs. Convolutional neural networks \citep{bruna2013spectral, Defferrard2016ConvolutionalNN, Kipf2017SemiSupervisedCW} borrow ideas from graph Laplacians \citep{cvetkovic1998spectra} for processing graphs signals.  \citet{Atwood2016DiffusionConvolutionalNN}, on the other hand, proposed a spatial approach relying on random walks. \citet{gilmer2017neural} proposed a general approach for learning on graphs via message passing between nodes of the graphs. All the other mentioned works can be interpreted as a restricted version of that approach.

\paragraph{Quantum Models}
There is a rich literature exploring quantum walks beginning with works of  \citet{ambainis2001one} and \citet{aharonov2001quantum}. A generalization of discrete walks for an arbitrary number of walkers was studied by \citet{rohde2011multi}. Subsequently, multiple works have developed graph kernels based on the quantum walks \citep{rossi2013continuous, bai2013quantum, bai2017quantum}. Quantum walks have also been shown to provide a model for universal computation\cite{childs_universal}. They have been explored for algorithmic applications \cite{Childs2003QuantumAF,Qiang_2012} and quantum simulation \cite{Berr_qwsim}. While there have been multiple proposals of quantum neural networks over the years \cite{gupta2001quantum, biamonte2017quantum}; there has been not much work done on quantum learning techniques for graphs. Our work derives from the recent work of \citet{Dernbach2018QuantumWI}, which proposed a quantum version of graph diffusion networks.

\section{Conclusion}
\label{sec:conc}
Quantum devices based on bosons are a prime candidate for future quantum. As such techniques which can directly leverage the behavior of bosons are important to explore. In this work we have tried to incorporate multi-particle bosonic walks on graphs. Unlike simple QWNN, this approach allows for learning significantly more powerful and complex graph diffusions. This benefit is clear across both regression and classification tasks. A future research direction would be to find ways to constrain the multiple walkers such that the simulation becomes more feasible.
%Another future direction of work is to directly incorporate quantum ideas into

\bibliography{mybib}
\bibliographystyle{abbrvnat}

\clearpage
\appendix

\setcounter{thm}{0}

\onecolumn

\section{Fock Space}
\label{apx:fock}
Fock basis is a construction for the state space of a variable or unknown number of identical particles from the Hilbert space of a single particle. If the identical particles are bosons, the $n$-particle states are vectors in the symmetric tensor product of $n$ single-particle Hilbert spaces \citep{fock_wiki}. In this section we provide a basic introduction to Fock spaces, and the corresponding creation and annihiliation operators. 

\begin{wrapfigure}{r}{4cm}
\vspace{-0.3cm}
\captionsetup{type=figure}
    \centering
    \includegraphics[width=\linewidth]{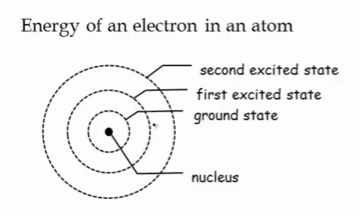}
    \caption{Configuration Space of an Atom \label{fig:hatom} }
\end{wrapfigure}

Consider a system which can accomodate particles in $M$ different states. Let $B = {k_i} i \in [0,1,..M]$ be an orthonormal basis of states in the the one-particle Hilbert space. For electronic states in an atom such as the one shown in Figure \ref{fig:hatom} these can correspond to the ground state shell , the first excited state or higher states. 

A Fock state is then defined as the state such that for each $i$, the state is an eigenstate of the particle number operator $ \widehat {N_{{\mathbf {k} }_{i}}}$ corresponding to the $i$-th elementary state $k_i$. The Fock bases of the state space of the system is then depicted by
$$\ket{n_0,n_1,n_2...n_M}$$
where $n_i$ denotes the number of particles in the state $i$. For the system in \ref{fig:hatom}  this refers to the number of electrons in different shells around the atom. Informally, the Fock bases is a counting basis i.e the bases of zero particle states, one particle states, two particle states, and so on

\paragraph{Creation and Annihilation Operators}
For the state defined earlier the  annihilation/creation operators are indexed by the state $i=1...M$ and operate as follows:
\begin{align*}
    a_i^\dagger \ket{n_0,n_1,..,n_i,..n_M} = n_i \ket{n_0,n_1,..,n_i-1,...,n_M} \\
    a_i \ket{n_0,n_1,..,n_i,..n_M} = n_i \ket{n_0,n_1,..,n_i+1,...,n_M} \\
\end{align*}

\section{Bosonic Walkers Example}
\label{apx:example}
Consider a pair of particles executing a quantum walk on a 1D grid.
The coin space of walkers then has two states $\ket{\uparrow}$ and $\ket{\downarrow}$. We assume that the shift operators act in a way such that a walker whose coin is up moves right, and a walker whose coin is down moves left.

Let the current state of the walkers after the action of the coin operator be: 
$$\ket{1,-1}\otimes\ket{\uparrow \downarrow}$$
It represents the first walker at position 1 with its coin state being up, while the second walker at position -1 with its coin state being down. From this state, the first particle can only move right and the second can only move left.

However if the particles are boson, then the aforementioned state does not exist in the Hilbert space of the system (as it is not symmetric with respect to the exchange operation). Instead the pair of bosonic walkers will be in a state $$ \frac{1}{\sqrt{2}}(\ket{1,-1}\otimes\ket{\uparrow \downarrow} + \ket{-1,1}\otimes\ket{\downarrow \uparrow})$$

From this state the first particle can move both left and right, a motion which is not possible for a non-bosonic walker. Note this was enforced by the symmetrization postulate. Such "nonlocal" correlations across non-interacting particle states is not possible in non-bosonic particles (or in classical case) and is a truly quantum phenomenon. %reminiscient of entanglement. %Nonlocality, quantum correlations, and violations of classical realism in the dynamics of two noninteracting quantum walkers

\section{Bosonic Walker Simulation}
\label{apx:bosonic}
We implement the simulation of quantum walkers as described in \cite{gamble2010twoparticle,Rigovacca_2016}. For simplicity, we shall limit the exposition to involve only 2 particles. The generalization to multiple walkers is straightforward.

A joint two boson state can be obtained from two single-particle states $\ket{\psi_1}, \ket{\psi_2}$ with the following symmetrization operation:

$$
\ket{\text{Sym}(\psi_1,\psi_2)} = \dfrac{\ket{\psi_1}\otimes\ket{\psi_2} + \ket{\psi_2}\otimes\ket{\psi_1}}{\sqrt{2(1+ |\bra{\psi_1}\ket{\psi_2}|^2)}}
$$

One should note that the aforementioned symmetrized form cannot represent all the possible states in the symmetric Hilbert space of two bosonic walkers. For example, in walks on a chain, the following state is a physically allowed one:
$$
\ket{\psi} = \dfrac{\ket{ij} + \ket{ji}}{\sqrt{2}} \otimes \dfrac{\ket{\uparrow \downarrow} + \ket{ \downarrow \uparrow}}{\sqrt{2}}
$$
where $i,j$ are indices of nodes in the chain, while $\downarrow,\uparrow$ represent the coin states. However, we shall limit ourselves to the states expressible in the aforementioned symmetrized form in this work.

Finally, during the measurement stage also, one needs to take into account the symmetrization requirement. This is done by considering a measurement involving the following projector operation::

$$
\Pi_{ij} = \frac{1}{2}(\ket{ij}\bra{ij} + \ket{ji}\bra{ji} 
$$
$$
Pr(\psi_t)[i,j] = \text{Tr}[\Pi_{ij} U_t\ket{\psi_0}\bra{\psi_0}U_t^\dagger]
$$

\section{Hyperparameter Details}
\label{apx:hyper}

\paragraph{Classification}
For the Enzyme and NCI1 experiment, we set the walk length to be 6 in both verison of the quantum network. The output neural net is a set2vec layer (for aggregation) followed by single layer. The feature and hidden layer dimensions are all set 64. In Mutag, the walk length is reduced to 4 and the layer size to 16. The GCN and DCNN are used as the input layer to a similar neural network i.e a set2vec layer followed by a hidden layer of size 64 (16 for Mutag).

\paragraph{Regression}
For QM7 we use quantum walk networks using a 4-step walk, followed by the set2vec layer with a hidden size of dimension 10. A similar setup is followed for DCNN and GCN models.

All models are trained with Adam optimizer with a learning rate of 1e-3.

\end{document}